\documentclass{ijuc}

\usepackage[prefix,norefpage,intoc]{nomencl}
\usepackage{mathrsfs,amsmath,amssymb,enumerate,amsfonts,epstopdf,subfigure,lineno}

\usepackage{graphicx}
\usepackage{url}
\usepackage{breqn}
\usepackage{color,soul}
\usepackage{float}
\usepackage{rotating}
\usepackage[export]{adjustbox}[2011/08/13]

\begin{document}

\title{Salt-n-pepper noise filtering using Cellular Automata}

\author{Dimitrios~Tourtounis, Nikolaos~Mitianoudis\email{nmitiano@ee.duth.gr} \and \\ Georgios Ch. Sirakoulis\email{gsirak@ee.duth.gr}
}

\institute{Democritus University of Thrace (DUTh)\\ Department of Electrical and Computer Engineering,\\University Campus DUTh, Xanthi 671 00, Greece}

\def\received{Received 2 January 2017; In final form XXXXXX}

\maketitle

\begin{abstract}
Cellular Automata (CA) have been considered one of the most pronounced parallel computational tools in the recent era of nature and bio-inspired computing. Taking advantage of their local connectivity, the simplicity of their design and their inherent parallelism, CA can be effectively applied to many image processing tasks. In this paper, a CA approach for efficient salt-n-pepper noise filtering in grayscale images is presented. Using a 2D Moore neighborhood, the classified ``noisy'' cells are corrected by averaging the non-noisy neighboring cells. While keeping the computational burden really low, the proposed approach succeeds in removing high-noise levels from various images and yields promising qualitative and quantitative results, compared to state-of-the-art techniques.
\end{abstract}


\section{Introduction}
\label{intro}

There are two most common types of noise in image processing: Gaussian noise and impulsive noise. Images are often corrupted by impulsive noise, which is caused by channel transmission errors, faulty memory locations in hardware or by malfunctioning pixels in camera sensors \cite{Gonzalez06}. Salt and pepper noise represents a special case of impulsive noise, where the corrupted image pixels can only take either the maximum or minimum values in the dynamic range. For this reason, salt and pepper noise normally appears either as black or white dots in an image. There are numerous techniques that attempt to efficiently restore an image corrupted by salt and pepper noise. Hitherto, median filtering has been the most common nonlinear filtering technique for removing this noise type. However, this is mainly effective for low noise densities. Moreover, the median filter applies the median operation to each pixel, regardless if it is noisy or not, which smears image details (such as edges and thin lines) \cite{Pitas90}. Thus, many improvements of the basic median filtering approach have been proposed. The Adaptive Median filter (AMF) is used to classify corrupted and uncorrupted pixels {performing} well at high noise densities. Although AMF showed promising results in removing noise, the window size in higher densities has to be large enough to remove the noise, resulting to increased computation complexity and {often blurred} restored images \cite{Hwang95}. Chan \emph{et al.} \cite{Chan05} proposed a two-phase solution. Firstly, an adaptive median filter is used to identify noisy pixels and secondly, image restoration is performed only to the previously selected noisy pixels using a specialized regularization method. This has shown to be very effective for high noise densities, nonetheless, the large window size increases the processing time.
Therefore, {Srinivasan and Ebenezer} \cite{Srinivasan07} recommended a new method, which corrects only corrupted pixels using the median value or its neighboring pixel value. The window size here remains equal to $3\times{3}$, thus reducing considerably the processing time. However, the edges of the restored image tend to appear less smooth and more pixelated. Another group of nonlinear filters has been proposed, including progressive switching median filter (PSMF) \cite{Wang99}, dynamic adaptive median filter (DAMF) \cite{Punyaban12b} and fuzzy based adaptive mean filter (FBAMF) \cite{Punyaban12}, which are adaptive, directional versions of the original median filter. A decision-based detail-preserving variational method (DPVM) for the removal of random-valued impulse noise was proposed, featuring an adaptive window type and size and a noise pixel annotation algorithm that guides the restoration algorithm to improve pixels accordingly \cite{Zhou12}.

There is also another group of image denoising algorithms, which are based on 2-D Cellular Automata (CA), that attempt to restore digital images corrupted by impulsive noise with the help of fuzzy logic theory \cite{Sahin14}. CA, although considered computational models of physical systems of discrete space and time \cite{Duff84}, have been successfully applied in image processing and computer vision \cite{Rosin14}. Lafe \cite{Lafe00} has also proposed CA methods, where information building blocks, called basis functions (or bases), can be generated from the evolving states of the CA, {namely} Cellular Automata Transforms (CAT) with direct application to image and video compression. More recently, it has also been shown by several researchers \cite{Andreadis,Ioannidis,Georgoulas2008,Zhao12a,Zhao12b,IoannidisSA13} that CA can be used to perform some standard image processing tasks to a high performance level, as well as in up-to-date computer vision fields, such as stereo vision \cite{Nalpantidis11,Nalpantidis11more}, image retrieval \cite{Konstantinidis}, medical image processing \cite{Gao2011,Hamamci,Dakua14}, image encryption \cite{Dogaru,Chatzichristofis,Jin2012,Wang2013,DelRey,Savvas}, image classification \cite{Espinola}, image coding \cite{Cappellari}, etc. For example, Rosin \cite{Rosin06,Rosin10} proposed training binary CA for noise filtering, thinning, and convex hull estimation. Another inherent advantage of CA is their parallelization capability that contributes to their performance increase. Furthermore, the CA approach is consistent with the modern notion of unified space-time. In computer science, space corresponds to memory and time to the processing unit. In CA, memory (CA cell state) and the processing unit (CA local rule) are inseparably related to a CA cell \cite{Sirakoulis15}. In terms of circuit design and layout, due to the ease of mask generation, silicon-area utilization, and the maximization of clock speed, CA are perhaps one of the most suitable computational structures for hardware realization \cite{Mardiris08}.

There were several recent applications of CAs on image edge detection. Uguz \emph{et al.} \cite{Uguz15} proposed a thresholding technique of edge detection based on fuzzy cellular automata transition rules enhanced using Particle Swarm Optimization. Hasanzadeh \emph{et al.} \cite{Hasanzadeh15} introduced a novel CA local rule with an adaptive neighborhood in order to produce the edge map of image. In contrast to common fixed neighborhood CAs, the proposed adaptive algorithm employs both von Neumann and Moore neighborhoods in an adaptive formulation. Finally, CAs have been also introduced into impulsive noise reduction in images. Selvapater and Hordijk \cite{wim} proposed a different modification of CA, such as a deterministic, random and mirrored CA to tackle the image noise filtering problem.  Preliminary CA are presented as a simplistic proof of concept that they could be an alternative to standard image noise filtering techniques\cite{jana}. A more enhanced CA based approach, in terms of the noise removal, was also presented \cite{abu2011}. A Cellular Automata Image Denoising (CAID) toolkit was introduced \cite{chinese} for the removal of salt and pepper noise in gray and color images. Sadeghi \emph{et al.} \cite{Sadeghi12} presented a hybrid method based on CA and fuzzy logic called Fuzzy Cellular Automata (FCA) in two steps. In the first step, noisy pixels are detected by CA, exploiting the local statistical information. In the second step, noisy pixels will be altered by FCA using the extracted statistical information. Finally, Sahin \emph{et al.} \cite{Sahin14} combine again two-dimensional CAs with the help of fuzzy logic theory. The algorithm employs a local fuzzy transition rule, which gives membership values to the corrupted neighboring pixels and assigns a next state value as a central pixel value.

A novelty of the proposed method is that it is applying CA to remove salt and pepper noise by altering only the pixels that have been corrupted, thus enhancing the performance of the applied CA-based method, while keeping the computational burden significantly low, along with the most advanced corresponding image processing techniques. In detail, the proposed algorithm is using a fixed $3\times{3}$ window size, to examine the 8 neighbors of the central pixel/CA cell, including the central pixel, in a Moore 2D CA neighborhood, which is applied to every pixel in the current image. Thus, the method's main advantages are that the CA is processing in real time and that the algorithm is self-adaptive, requiring only a rough estimate of noise percentage to be defined. Another advantage of this algorithm is that it requires significantly lower computational time compared to other algorithms and the results even in very high noise densities, such as 80\% or 90\%, are satisfactory, giving smoother restored images than other methods. The proposed method's maximum possible complexity scales linearly with the noise level, which provides a speed benefit compared to many other approaches. On top of all these, the inherent parallelism of CA enables the straightforward hardware implementation of the proposed really simple CA-based method without any hardware overhead. As a result, the simplicity of the proposed method, its minimal complexity and its evolution through time when combined with the inherent parallelism of the CA approach result in a quite efficient filtering procedure. In this study, we compare with a family of adaptive median filters as well as other well known denoising techniques which the proposed method outperforms in terms of Peak Signal-to-Noise Ratio (PSNR) and Structural Similarity (SSIM) \cite{Wang04}. A similar trend appears when the proposed approach is compared in terms of PSNR and SSIM with all the corresponding CA based techniques dealing with salt and pepper noise removal, as encountered in modern literatur, to the best of our knowledge, and described earlier.

This paper is organized as follows. In Section \ref{CA}, we introduce the basic principles of the  CA computational tool. Section \ref{method} describes the proposed method and the necessary steps to implement the algorithm, while in Section \ref{Results}, we present the results of the proposed method and its comparison among the other methods that already exist. This comparison is based on PSNR and SSIM values. Experiments show that the proposed method performs better than the other {existing} methods. Finally, Section \ref{Conclusions} concludes the paper.

\section{Cellular Automata Principles}
\label{CA}
Cellular Automata (CA) are a very elegant computing model, which dates back to John von Neumann \cite{vonNeumann52}. CA decompose problems into a field of cells and a local rule, which defines the new state of a cell, depending on its neighbors' states. All cells can operate in parallel, since each cell can independently update its own state. Hence, CA can capture the essential features of systems, where global behavior arises from the collective effect of simple components, which interact locally. In addition, the model is massively parallel and ideal for hardware implementation. In general, a CA requires \cite{Chopard98}:

\begin{enumerate}
\item a regular lattice of cells covering a portion of a $d$-dimensional space;

\item a set $\mathbf{C}(\vec{r},t)=\{C_1(\vec{r},t),C_2(\vec{r},t),\dots,C_m(\vec{r},t)\}$ of variables attached to each site $\vec{r}$ of the lattice giving the local state of each cell at the time $t = 0, 1, 2, \dots $ ;

\item a rule $\mathbf{R}= \{R_{1}, R_{2}, \dots , R_{m}\}$, which specifies the time evolution of the states $\mathbf{C}(\vec{r},t)$ in the following way: ${C}_j(\vec{r},t+1)=R_j(\mathbf{C}(\vec{r},t),\mathbf{C}(\vec{r}+\vec{\delta_1},t),\mathbf{C}(\vec{r}+\vec{\delta_2},t),\dots,\mathbf{C}(\vec{r}+\vec{\delta_q},t))$, where $\vec{r}+\vec{\delta_k}$ designate the cells belonging to a given neighbourhood of cell $\vec{r}$.
\end{enumerate}

{In the above definition, the rule $\mathbf{R}$ is identical for all sites and is applied simultaneously to each of them, leading to synchronous dynamics. It is important to notice that the rule is homogeneous, i.e. it does not depend explicitly on the cell position $\vec{r}$. However, spatial (or even temporal) inhomogeneities can be introduced by having some $C(\vec{r})$ systematically at 1, in some given locations of the lattice, to mark particular cells for which a different rule applies. Furthermore, in the above definition, the new state at time $t+1$ is only a function of the previous state at time $t$. It is sometimes necessary to have a longer memory and introduce a dependence on the states at time $t-1, t-2, \dots, t-k$. Such a situation is already included in the definition, if one keeps a copy of previous states in the current state.}

The neighbourhood of a cell $\vec{r}$ is the spatial region in which a cell needs to search in its vicinity. In principle, there is no restriction on the size of the neighbourhood, except that it is the same for all cells. However, in practice, it is often made up of adjacent cells only. For 2-D CA, two neighbourhoods are commonly considered: The von Neumann neighbourhood, which consists of a central cell and its four geographical neighbours north, west, south and east. The Moore neighbourhood is a super set containing second nearest neighbours, i.e. northeast, northwest, southeast and southwest, giving a total of nine cells. In practice, when simulating a given CA rule, it is impossible to deal with an infinite lattice. The system must be finite and have boundaries. Clearly, a site belonging to the lattice boundary does not have the same neighbourhood as other internal sites. In order to define the behaviour of these sites, the neighbourhood is extending for the sites at the boundary {leading} to various types of boundary conditions, such as periodic (or cyclic), fixed, adiabatic or reflection.

\section{Proposed Denoising Method}
\label{method}
In this paper, a novel method based on CA is applied to remove impulsive noise from gray-scale images. The proposed method was inspired from the Segmentation Matching Factor \cite{Bovik00}, where each pixel is replaced by the median of its neighborhood values. Nevertheless, the approach presented here is somehow different. 
We consider a 2-D image which is divided into a matrix of identical square CA cells, with side length $a$ and is represented by a CA. For matters of simplicity, we consider each CA cell an image pixel; so the number of spatial dimensions of the CA array is $n=2$, while the widths of the two sides of the CA array are taken to be equal, i.e. $w_{1}=w_2$. We also assume zero boundary conditions for the CA. In the case of $C_{(i_0,j_0)}$, the under study pixel at position $(i_0,j_0)$, the state of the corresponding CA cell is made to take $256$ discrete values as follows:

\begin{equation}
{C^t}_{(i_0,j_0)}\in \{0,\dots ,255\}
\label{eqtn1}
\end{equation}

This is due to the assumption that the intensity of each pixel is represented by 8-bit gray-scale accuracy. Furthermore, the Moore ($M$) neighborhood ($N$) for the range $r$ of a CA cell $C_{(i_0,j_0)}\ $can be defined by the following equation:

\begin{equation}
\label{GrindEQ__2_}
{N\left(i_0,j_0\right)}^M=\{\left(i,j\right):\left|i-i_0\right|\le r,\ \left|j-j_0\right|\le r\}
\end{equation}

In our case, range $r$ equals to 1, resulting in a fixed neighborhood size of $3\times 3$, which is used for the whole image. As mentioned before, two thresholds are considered for the CA state values, i.e. 
$min_{state}=0$ and
$max_{state}=255$. In general, the local 2D rule for the proposed CA is given as follows:

\small
\begin{equation}
\label{GrindEQ__3_}
{C^{t+1}}_{(i,j)}=\left\{ \begin{array}{ll}
{C^t}_{\left(i,j\right)},&{\rm if } \  {min_{state} <C^t}_{\left(i,j\right)}< max_{state} \\
{{Cnew}^t}_{(i,j)}\ ,&{\rm if }\  C^t_{\left(i,j\right)}=min_{state} \\&{\rm or }{\ C}^t_{\left(i,j\right)}\ =max_{state}\  \end{array}
\right.\
\end{equation}
\normalsize

In (\ref{GrindEQ__3_}), the new value ${C^{t+1}}_{(i,j)}$ of CA cell ${C^t}_{(i,j)}$ is calculated as a local 2D sub-rule described by (\ref{eq4}) as found below: 

\begin{equation}
\resizebox{.99\hsize}{!}{$Cnew^t_{(i,j)}=\left\{ \begin{array}{ll}
{\rm mean}_{-r\le i,j\le r}(C^t_{i,j}),&\ {\rm if}\ \ \forall \ C^t_{(i\pm r,j\pm r)}\in N,\ C^t_{(i\pm r,j\pm r)}\ne min_{state} {\rm or}\ C^t_{(i\pm r,j\pm r)}\ne max_{state},\ {\rm where}\ r=1 \\
{\rm mean}_{-r\le i,j\le r}(C^t_{(i,j)}),&\ {\rm if}\ \ \forall \ C^t_{(i\pm r,j\pm r)}\in N,\ \exists C^t_{(i\pm r,j\pm r)}= min_{state}\ {\rm and}\ C^t_{(i\pm r,j\pm r)}= max_{state},\ {\rm where}\ r=1 \\
max_{state},&\  {\rm if}\ \forall \ C^t_{(i\pm r,j\pm r)},\ C^t_{(i\pm r,j\pm r)}= min_{state}\ {\rm or}\ C^t_{(i\pm r,j\pm r)}= max_{state}\  \end{array}
\right. $}
\label{eq4}
\end{equation}

As a result, in the proposed CA the requested detection of noisy and noisy-free pixels is given by the corresponding CA rules, as previously described, by checking the value of the CA cell itself and the values of the corresponding Moore neighborhoods. For the sake of simplicity, we clarify that if the value of the under study CA cell in each neighborhood is defined by the aforementioned thresholds, this implies that the corresponding CA cell is defined as a ``noisy'' one. This is due to the {salt-n-pepper} noise that influences the CA cell, by replacing its state by either a minimum or a maximum value in the range of the CA cell discrete states. The proposed rule replaces the noisy pixels with a mean of the neighbouring cells that are not in a min\_{state} or a max\_{state}. In the case that the CA cell state is not equal to any of the threshold values, then the CA cell is not considered a noisy one and consequently, its state will be kept unchanged. Otherwise, the CA evolution subrules should be applied and the CA cell state has to be estimated accordingly, since it is considered a noisy/corrupted one. The whole CA evolves for a finite number of iterations, depending on the level of noise. As a rule of thumb, if the level of noise is $n\%$, the CA iterates for $n/10+1$ iterations.

Recapitulating, the pseudocode of the proposed CA algorithm shows the steps followed in the proposed method.

\textbf{\textit{Pseudocode of the proposed CA Algorithm}}\\
\textbf{Step 1:} Read the original image $I(x,y)$.\\
\textbf{Step 2:} \textit{If} $I(x,y)$ is in RGB, then convert to grayscale, or work independently on each color channel.\\
\textbf{Step 3:} Assume a 2-D window of size $3\times 3$, which scans the image $I(x,y)$.\\
\textbf{Step 4:} Let $C_{i,j}$ represent the central pixel of a 2D Moore's neighborhood in the CA. \\
\textbf{Step 5:} Create a vector $B$, which has dimensions $8\times 1$. The pixel values inside the window, excluding the central pixel, are sorted in this matrix. These values are arranged in ascending order.\\
\textbf{Step 6:} Let $B_{min}$ and $B_{max}$ represent the minimum and maximum pixel values.\\
\textbf{Step 7:} \textit{If} $0<C_{i,j}<255$, $C_{i,j}$ is an uncorrupted pixel and it will be kept unchanged.\\
\textbf{Step 8:} \textit{If} $C_{i,j}$ is a noisy pixel (i.e. $C_{i,j}=0$ $\vee$ $C_{i,j}=255$) \textit{then}

\textit{\textbf{Case 1:}} \textit{If} $B_{min}=0$ $\wedge$ $B_{max}=255$ \textit{then} \\
  $C_{i,j}$=mean $(B)$ without $B_{min}=0$ and $B_{max}=255$ \\
  \textit{endif}

\textit{\textbf{Case 2:}} \textit{If} (all elements of $B =0$ $\vee$ $B  =255$) \textit{then} \\
  $C_{i,j}=255$ \\
  \textit{endif}

\textit{\textbf{Case 3:}} \textit{If} $B_{min}>0$ $\wedge$ $B_{max}<255$ \textit{then} \\
$C_{i,j}$=mean $B$ \\
  \textit{endif}

\noindent\textbf{Step 9:} Repeat steps (6)-(8) for all the pixels of input image $I(x,y)$ for $n/10+1$ iterations ($n\%$ is the level of noise).

In the proposed method, during step (8) we are testing 3 cases, where the central pixel is a noisy one. The key idea of our algorithm among other methods is, that we calculate the mean value of the selected window by first removing the maximum and minimum values in the dynamic range (0,255) if they exist in the neighborhood. This provides less abrupt edge transitions, leading to smoother edge preservation for noise densities varying from $10\%-90\%$. Finally, the computational complexity for the $N\times N$ 2D CA is $O(N^2)$.

\begin{table}[htbp]
\caption{Restoration results in terms of PSNR (dB) (left) and SSIM (right) for different rates of impulsive noise density for the $256\times 256$ Lena image.}
\begin{center}
\begin{adjustbox}{center}
\resizebox{1.3\hsize}{!}{
\begin{tabular}{|c|c|c|c|c|c|c|c|c|c|c|c|c|c|c|c|} \hline
Noise Ratio & \multicolumn{2}{|c|}{AMF \cite{Hwang95}} & \multicolumn{2}{c|}{BDND \cite{Ng06}} & \multicolumn{2}{c|}{MBUTMF \cite{Esakkirajan11}} & \multicolumn{2}{c|}{DWMF \cite{Dong07}} & \multicolumn{2}{c|}{MDWMF \cite{Lu12}} & \multicolumn{2}{c|}{Li \textit{et al.} \cite{Li14}} & \multicolumn{2}{c|}{Proposed Method} \\ \hline
10\% & 35.2 & 0.9797 & 39.1 & 0.991 & 40.2 & 0.9921 & 33.3 & 0.9701 & 37.0 & 0.9836 & 39.5 & 0.9914 & \textbf{41.2} & \textbf{0.9929} \\ \hline
20\% & 33.2 & 0.9674 & 34.7 & 0.9772 & 36.3 & 0.9823 & 30 & 0.9546 & 33.4 & 0.9641 & 36.3 & 0.982 & \textbf{37.9} & \textbf{0.9838} \\ \hline
30\% & 30.7 & 0.9426 & 29.5 & 0.9269 & 33.7 & 0.9670 & 28.3 & 0.9307 & 31.3 & 0.9377 & 33.9 & 0.9689 & \textbf{34.7} & \textbf{0.9748} \\ \hline
40\% & 28.5 & 0.9083 & 25.9 & 0.8525 & 31.5 & 0.9480 & 26.7 & 0.8704 & 29.6 & 0.9102 & 32.1 & 0.9524 & \textbf{33.0} & \textbf{0.9619} \\ \hline
50\% & 26.6 & 0.8667 & 22.4 & 0.7256 & 29.6 & 0.9169 & 24.9 & 0.8096 & 28.1 & 0.8752 & 30.1 & 0.9269 & \textbf{31.3} & \textbf{0.9484} \\ \hline
60\% & 24.5 & 0.8048 & 20.1 & 0.6075 & 26.9 & 0.8434 & 23.4 & 0.7524 & 26.6 & 0.8306 & 27.8 & 0.8814 & \textbf{29.8} & \textbf{0.927} \\ \hline
70\% & 22.7 & 0.7271 & 18.7 & 0.4939 & 23.7 & 0.6904 & 20.7 & 0.6127 & 25.1 & 0.7569 & 26.7 & 0.8464 & \textbf{28.1} & \textbf{0.9007} \\ \hline
80\% & 20.3 & 0.6099 & 17.9 & 0.4468 & 19.8 & 0.4423 & 18.2 & 0.3054 & 23.5 & 0.6296 & 25.1 & 0.7889 & \textbf{26.2} & \textbf{0.8612} \\ \hline
90\% & 17.0 & 0.4457 & 15.3 & 0.3853 & 15.7 & 0.2063 & 12.9 & 0.0679 & 21.0 & 0.4744 & 23.3 & 0.6985 & \textbf{23.7} & \textbf{0.7904} \\ \hline
\end{tabular}}
\end{adjustbox}
\end{center}
\label{Table1}
\end{table}

\begin{figure*}[!t]
\centering
\includegraphics[scale=.3,center]{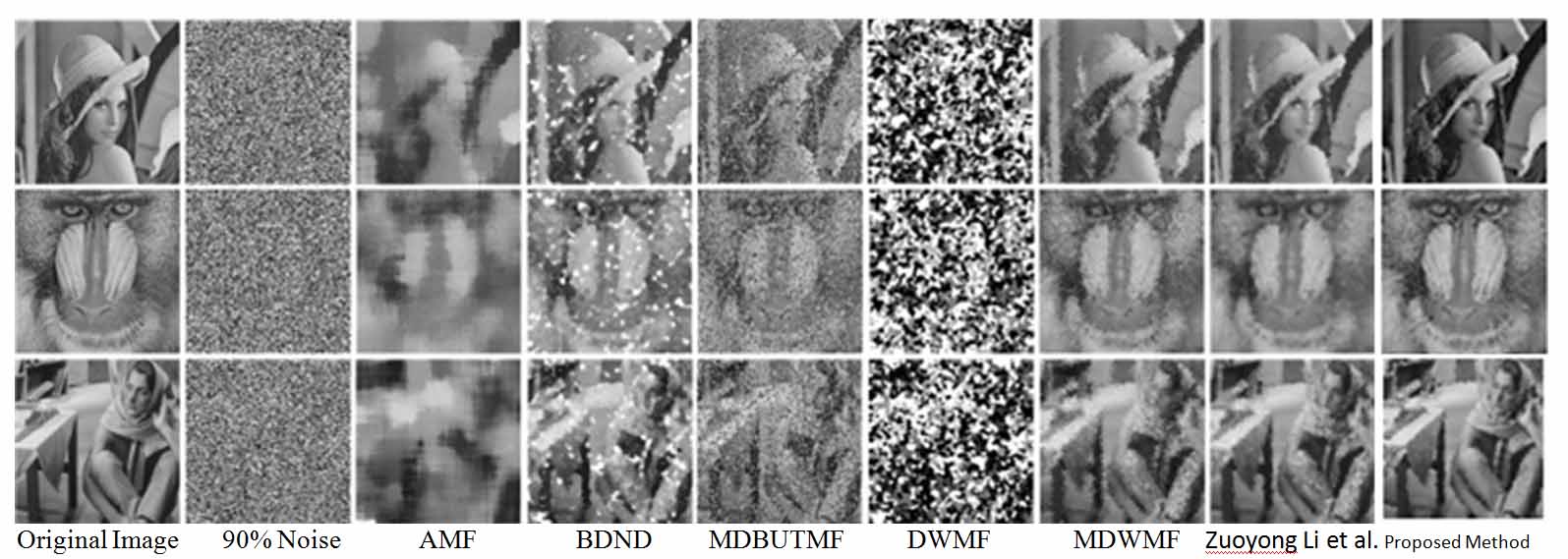}
\caption{Restored images using different filters{, namely AMF \cite{Hwang95}, BDND \cite{Ng06}, MDBUTMF \cite{Esakkirajan11}, DWMF \cite{Dong07}, MDWMF \cite{Lu12}, Zuoyong Li \textit{et al.} \cite{Li14}, and the proposed method for $90\%$ of salt and pepper noise for different  $256\times 256$ pixel images like Lena, Baboon and Barbara.}}
\label{Fig7}
\end{figure*}

\begin{figure*}[!t]
\centering
\includegraphics[scale=.4,center]{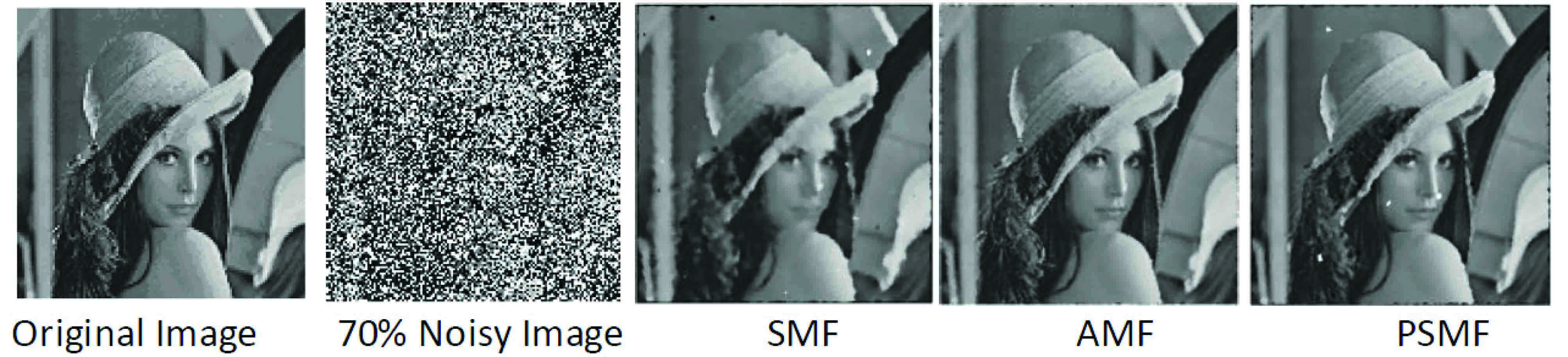}\\
\includegraphics[scale=.4,center]{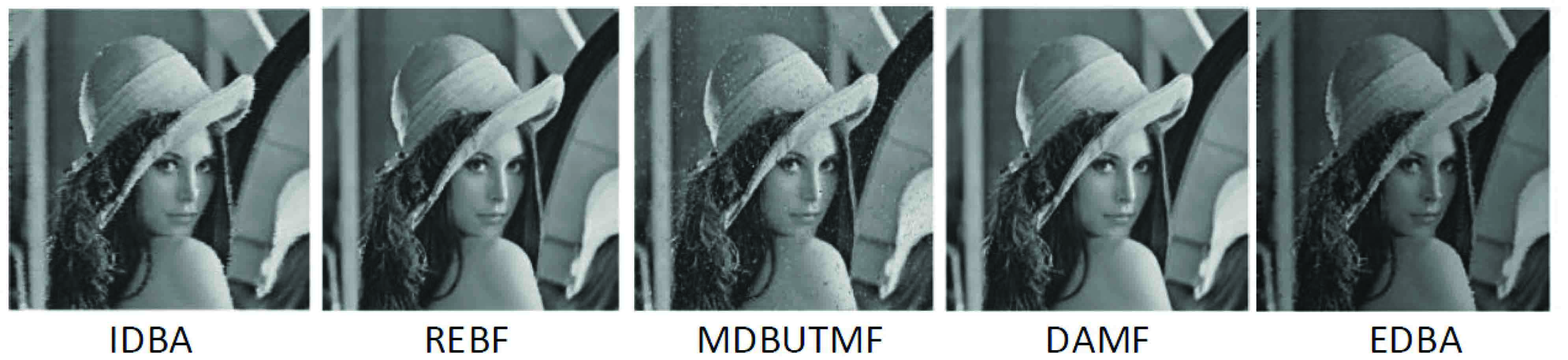}\\
\includegraphics[scale=.48,center]{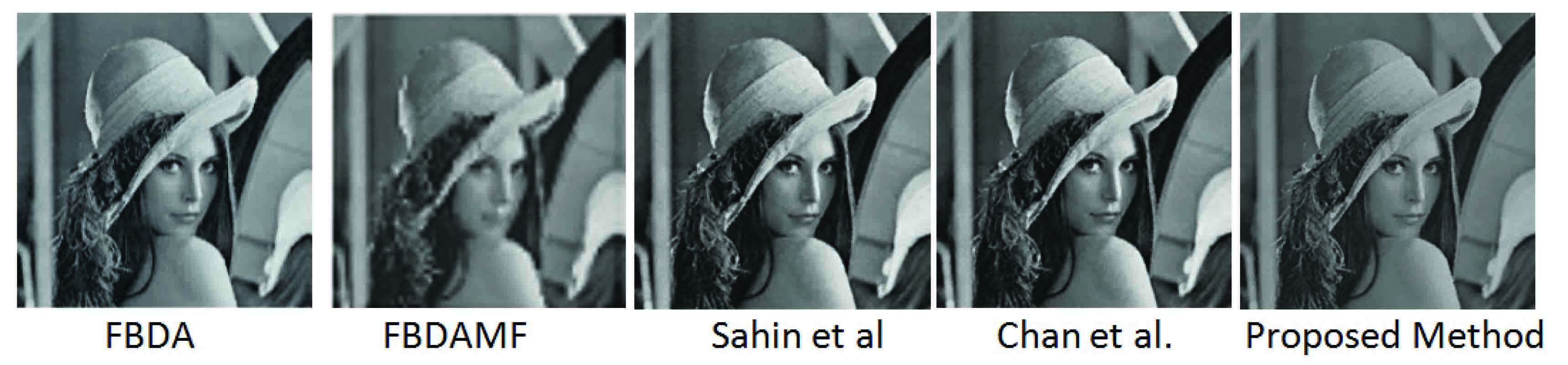}
\caption{Restored images using different filters, namely SMF \cite{Bovik00}, AMF \cite{Hwang95}, PSMF \cite{Wang99}, IDBA \cite{Nair08}, REBF \cite{Vijaykumar08}, MDBUTMF \cite{Esakkirajan11}, DAMF \cite{Punyaban12b}, EDBA \cite{Srinivasan07}, FBDA \cite{Nair10}, FBDAMF \cite{Punyaban12}, Sahin \textit{et al.} \cite{Sahin14}, Chan \textit{et al.} \cite{Chan05} and the proposed method for $70\%$ of salt and pepper noise for the $512\times 512$ Lena image.}
\label{Fig4}
\end{figure*}

\begin{figure*}[!t]
\centering
\includegraphics[scale=.4,center]{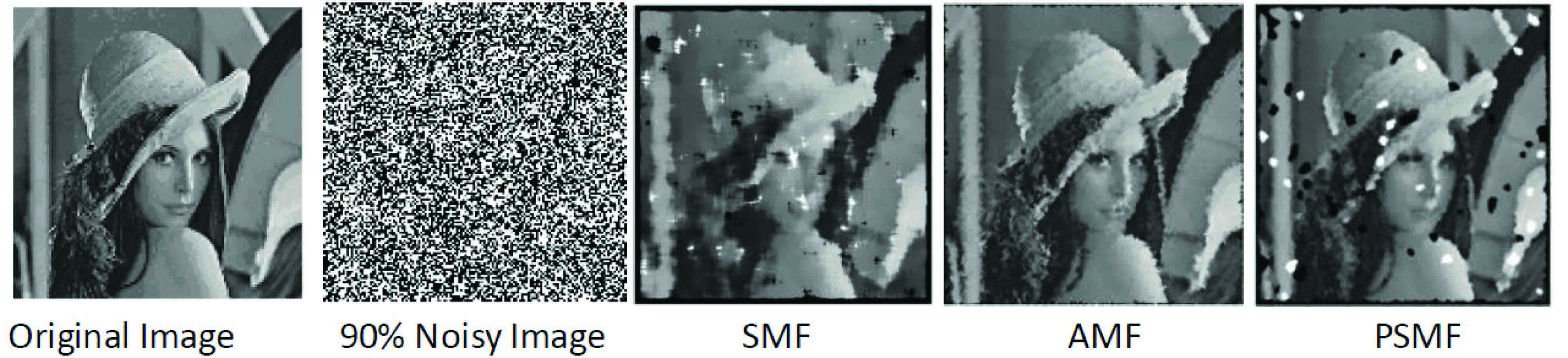}\\
\includegraphics[scale=.4,center]{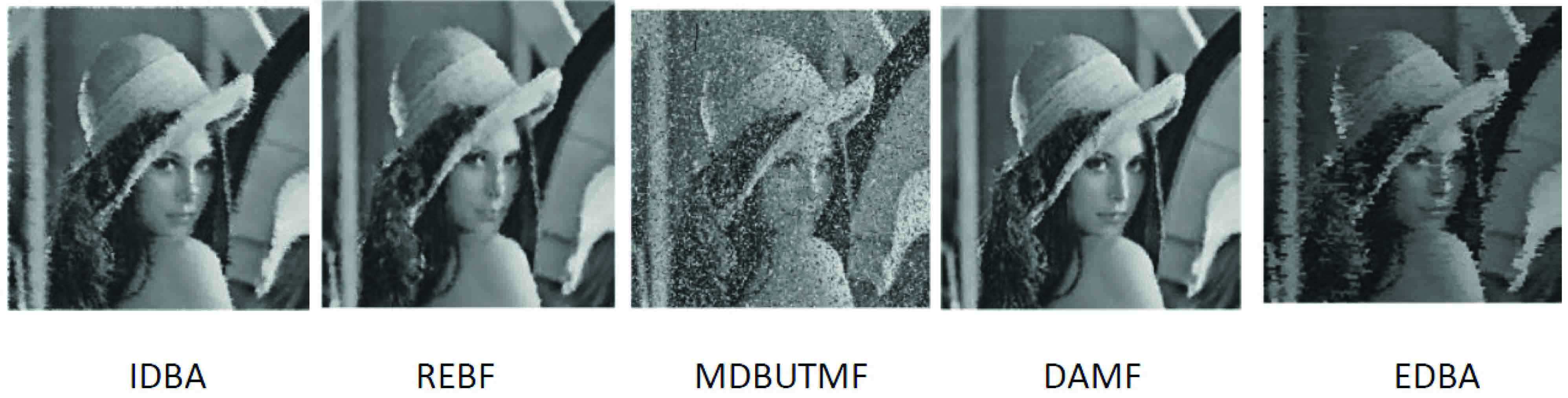}\\
\includegraphics[scale=.4,center]{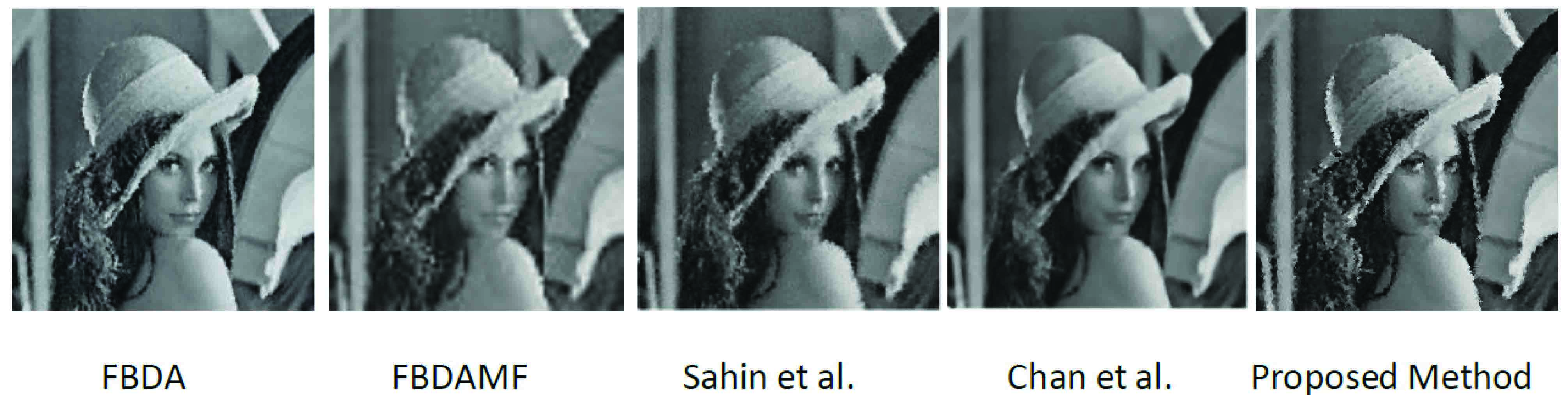}
\caption{Restored images using different filters for $90\%$ of salt and pepper noise for the $512\times 512$ Lena image.}
\label{Fig5}
\end{figure*}

\begin{table}[htbp]
\caption{Restoration results in terms of PSNR (dB) for different rates of impulsive noise density for the $512\times 512$ Lena image.}
\begin{center}

\begin{adjustbox}{center}
\resizebox{1.3\hsize}{!}{
\begin{tabular}{|c|c|c|c|c|c|c|c|c|c|c|c|c|c|c|c|c|} \hline
{Noise} & {SMF} & {PSMF} & {AMF} & {IDBA} & {MDWMF} & {Fuzzy} & {EDBA} & {MDBUTMF} & {Chan} & {Sahin} & {FBAMF} & {FBDA} & {REBF} & {DAMF} & {Pattnaik} & {Proposed} \\
{Ratio} & {\cite{Bovik00}} & \cite{Wang99} & \cite{Hwang95} & \cite{Nair08} & \cite{Lu12} & \cite{Thirilogasundari12} & \cite{Srinivasan07} & \cite{Esakkirajan11} & \textit{et al.} \cite{Chan05} & \textit{et al.} \cite{Sahin14} & \cite{Punyaban12} & \cite{Nair10} & \cite{Vijaykumar08} & \cite{Punyaban12b} & \textit{et al.} \cite{Pattnaik12} & {Method} \\ \hline
10\% & 36.12 & 37.01 & 38.76 & 39.59 & 41.45 & 38.38 & 38.43 & 44.32 & 42.6 & 40.7 & 44.02 & 39.88 & 39.93 & 44.47 & 41.87 & \textbf{47.6795} \\ \hline
20\% & 33.42 & 33.45 & 35.01 & 36.92 & 38.22 & 37.47 & 37.36 & 40.3 & 39.3 & 37.1 & 40.51 & 37.83 & 38.49 & 40.3 & 38 & \textbf{43.9804} \\ \hline
30\% & 31.36 & 30.86 & 32.26 & 34.61 & 35.97 & 36.02 & 35.92 & 37.99 & 37.0 & 34.9 & 38.24 & 36.1 & 36.97 & 37.99 & 35.75 & \textbf{41.3465} \\ \hline
40\% & 29.88 & 27.56 & 30.09 & 32.74 & 34.1 & 34.54 & 34.12 & 35.95 & 34.3 & 33.2 & 36.44 & 34.36 & 35.51 & 35.95 & 33.83 & \textbf{39.0329} \\ \hline
50\% & 28.54 & 26.35 & 28.49 & 30.91 & 32.69 & 33.09 & 32.21 & 34.42 & 31.8 & 31.8 & 35.0 & 33.08 & 33.97 & 34.42 & 32.1 & \textbf{37.1154} \\ \hline
60\% & 26.76 & 24.55 & 26.61 & 29.38 & 31.21 & 31.73 & 30.43 & 33.04 & 30.8 & 30.5 & 33.34 & 31.75 & 32.43 & 33.04 & 30.62 & \textbf{35.1941} \\ \hline
70\% & 24.47 & 23.04 & 24.25 & 27.99 & 29.72 & 30.22 & 28.62 & 31.13 & 29.7 & 29.2 & 31.38 & 30.07 & 30.75 & 31.13 & 28.86 & \textbf{33.1756} \\ \hline
80\% & 19.52 & 20.23 & 23.23 & 25.89 & 27.94 & 28.4 & 26.23 & 28.71 & 27.5 & 27.2 & 29.51 & 28.53 & 28.92 & 28.71 & 26.93 & \textbf{31.0194} \\ \hline
90\% & 8.8 & 15.9 & 20.71 & 22.8 & 25.5 & 24.04 & 23.94 & 26.43 & 25.4 & 25.7 & 26.91 & 26.68 & 25.21 & 26.43 & 24.61 & \textbf{27.9889} \\ \hline
\end{tabular}}
\end{adjustbox}
\end{center}
\label{Table1plus}
\end{table}

\begin{table}[htbp]
\caption{Comparisons of restoration results in SSIM for different rates of impulsive noise density for Lena image with resolution $512\times 512$.}
\begin{center}
\begin{tabular}{|c|c|c|c|c|c|c|c|} \hline
Noise & SMF & AMF & EDBA & IDBA & BDND & FBDA & Proposed \\
Ratio & {\cite{Bovik00}} & \cite{Hwang95} & \cite{Srinivasan07} & \cite{Nair08} & \cite{Ng06} & \cite{Nair10} & Method \\ \hline
\multicolumn{8}{|c|}{SSIM values} \\ \hline
10\% & 0.9931 & 0.9974 & 0.9951 & 0.9978 & 0.9989 & 0.9979 & \textbf{0.9994} \\ \hline
20\% & 0.9812 & 0.9939 & 0.9914 & 0.9963 & 0.9981 & 0.9971 & \textbf{0.9986} \\ \hline
30\% & 0.9718 & 0.9886 & 0.9879 & 0.9941 & 0.9962 & 0.9963 & \textbf{0.9973 }\\ \hline
40\% & 0.9614 & 0.9825 & 0.9825 & 0.9901 & 0.9933 & 0.9948 & \textbf{0.9954} \\ \hline
50\% & 0.9381 & 0.9738 & 0.9755 & 0.9843 & 0.9893 & 0.9899 & \textbf{0.9928} \\ \hline
60\% & 0.9155 & 0.9636 & 0.9655 & 0.9749 & 0.9831 & 0.9842 & \textbf{0.9885} \\ \hline
70\% & 0.8646 & 0.9471 & 0.9483 & 0.9638 & 0.9766 & 0.9974 & \textbf{0.98} \\ \hline
80\% & 0.7939 & 0.9209 & 0.9154 & 0.9491 & 0.9697 & 0.9593 &\textbf{ 0.9642} \\ \hline
90\% & 0.6388 & 0.8637 & 0.8132 & 0.9152 & 0.9546 & 0.9325 & \textbf{0.9165 }\\ \hline
\end{tabular}
\end{center}
\label{Table5}
\end{table}

\begin{table}[htbp]
\caption{Restoration results in terms of PSNR (dB) (left) and SSIM (right) for different rates of impulsive noise density for the $256\times 256$ Baboon image.}
\begin{center}
\begin{adjustbox}{center}
\resizebox{1.3\hsize}{!}{
\begin{tabular}{|c|c|c|c|c|c|c|c|c|c|c|c|c|c|c|c|} \hline
Noise Ratio & \multicolumn{2}{|c|}{AMF \cite{Hwang95}} & \multicolumn{2}{c|}{BDND \cite{Ng06}} & \multicolumn{2}{c|}{MBUTMF \cite{Esakkirajan11}} & \multicolumn{2}{c|}{DWMF \cite{Dong07}} & \multicolumn{2}{c|}{MDWMF \cite{Lu12}} & \multicolumn{2}{c|}{Li \textit{et al.} \cite{Li14}} & \multicolumn{2}{c|}{Proposed Method} \\ \hline
10\% & 29.6 & 0.9269 & 33.9 & 0.9725 & 34.3 & 0.9747 & 25.8 & 0.8216 & 32.1 & 0.9594 & 34.4 & 0.9754 & \textbf{34.43} & \textbf{0.9757} \\ \hline
20\% & 28.8 & 0.9118 & 30.2 & 0.9372 & 30.9 & 0.9447 & 25.1 & 0.7866 & 28.9 & 0.9150 & 31.1 & 0.9472 & \textbf{31.20} & \textbf{0.9458} \\ \hline
30\% & 26.9 & 0.8581 & 26.7 & 0.8709 & 28.8 & 0.9076 & 24.2 & 0.7416 & 26.9 & 0.8614 & 29.1 & 0.9117 & \textbf{ 29.28} & \textbf{0.9131} \\ \hline
40\% & 25.4 & 0.7989 & 23.5 & 0.7714 & 27.2 & 0.8659 & 23.1 & 0.6293 & 25.3 & 0.8014 & 27.6 & 0.8718 & \textbf{27.76} & \textbf{0.8773} \\ \hline
50\% & 24.4 & 0.7326 & 21.2 & 0.6532 & 25.9 & 0.8191 & 22.2 & 0.4933 & 24.2 & 0.7433 & 26.3 & 0.828 & \textbf{26.50} & \textbf{0.8321} \\ \hline
60\% & 23.1 & 0.6407 & 19.3 & 0.5118 & 24.2 & 0.7324 & 21.0 & 0.4485 & 22.9 & 0.6697 & 24.5 & 0.7459 & \textbf{ 25.24} & \textbf{0.7747} \\ \hline
70\% & 22.0 & 0.5535 & 18.3 & 0.4143 & 22.1 & 0.6120 & 17.7 & 0.3607 & 21.8 & 0.5793 & 23.6 & 0.6478 & \textbf{ 23.97} & \textbf{0.7056} \\ \hline
80\% & 20.8 & 0.4467 & 17.5 & 0.3347 & 19.4 & 0.4368 & 13.2 & 0.2056 & 20.2 & 0.4393 & 22.5 & 0.5603 & \textbf{ 22.68} & \textbf{0.6174} \\ \hline
90\% & 19.1 & 0.3271 & 15.2 & 0.2396 & 16.2 & 0.2108 & 8.5 & 0.05106 & 19.2 & 0.3128 & 21.3 & 0.4068 & \textbf{21.32} & \textbf{0.4822} \\ \hline
\end{tabular}}
\end{adjustbox}
\end{center}
\label{Table2}
\end{table}

\begin{table}[htbp]
\caption{Restoration results in terms of PSNR (dB) (left) and SSIM  (right) for different rates of impulsive noise density for the $256\times 256$ Barbara image.}
\begin{center}
\begin{adjustbox}{center}
\resizebox{1.3\hsize}{!}{
\begin{tabular}{|c|c|c|c|c|c|c|c|c|c|c|c|c|c|c|c|} \hline
Noise Ratio & \multicolumn{2}{|c|}{AMF \cite{Hwang95}} & \multicolumn{2}{c|}{BDND \cite{Ng06}} & \multicolumn{2}{c|}{MBUTMF \cite{Esakkirajan11}} & \multicolumn{2}{c|}{DWMF \cite{Dong07}} & \multicolumn{2}{c|}{MDWMF \cite{Lu12}} & \multicolumn{2}{c|}{Li \textit{et al.} \cite{Li14}} & \multicolumn{2}{c|}{Proposed Method} \\ \hline
10\% & 30.5 & 0.9599 & 31.3 & 0.9699 & 31.7 & 0.9730 & 23.4 & 0.8051 & 30.6 & 0.9619 & 32.2 & 0.9739 & \textbf{39.3} & \textbf{0.9883} \\ \hline
20\% & 28.4 & 0.9378 & 27.7 & 0.9328 & 28.3 & 0.9405 & 22.9 & 0.7421 & 27.1 & 0.9163 & 29.1 & 0.9455 & \textbf{35.5} & \textbf{0.975} \\ \hline
30\% & 26.7& 0.9037 & 25.4 & 0.8791 & 26.4 & 0.9040 & 22.4 & 0.7186 & 25.3 & 0.8671 & 27.4 & 0.9139 & \textbf{33.4} & \textbf{0.9588} \\ \hline
40\% & 25.1 & 0.8566 & 22.8 & 0.7954 & 25.0 & 0.8637 & 21.8 & 0.6346 & 23.8 & 0.8121 & 25.9 & 0.8789 & \textbf{32.0} & \textbf{0.9406} \\ \hline
50\% & 23.6 & 0.8002 & 20.4 & 0.6705 & 23.7 & 0.8079 & 21.2 & 0.6111 & 22.3 & 0.7409 & 24.8 & 0.8345 & \textbf{30.4} & \textbf{0.918} \\ \hline
60\% & 22.0 & 0.7205 & 18.5 & 0.5581 & 22.3 & 0.7215 & 19.9 & 0.5666 & 21.2 & 0.6765 & 23.9 & 0.7895 & \textbf{28.7} & \textbf{0.8851} \\ \hline
70\% & 20.4 & 0.6193 & 17.5 & 0.4711 & 20.3 & 0.5822 & 16.9 & 0.4593 & 19.8 & 0.5652 & 22.9 & 0.7339 & \textbf{27.1} & \textbf{0.8465} \\ \hline
80\% & 18.4 & 0.4732 & 16.8 & 0.3988 & 17.7 & 0.3825 & 12.3 & 0.2443 & 18.6 & 0.4419 & 21.8 & 0.6544 & \textbf{25.5} & \textbf{0.7879} \\ \hline
90\% & 15.1 & 0.2488 & 14.4 & 0.3202 & 14.6 & 0.1922 & 8.4 & 0.0695 & 17.2 & 0.3160 & 20.4 & 0.5503 & \textbf{23.1} & \textbf{0.6913} \\ \hline
\end{tabular}}
\end{adjustbox}
\end{center}
\label{Table3}
\end{table}

\section{Experimental Results}
\label{Results}
In this section, the performance of our algorithm is tested on different grayscale images. The experimental images are common natural images used in image processing, {such as} Lena and Bridge images, at $256\times 256$ and $512\times 512$ pixel resolution, with varying percentage of salt and pepper noise. It is valid to compare denoising performance on the same image at different resolutions, since denoising is much more difficult at lower resolutions. We experimented with noise levels ranging from 10\% to 90\% with an increase of 10\%. To evaluate the restoration performance of the traditional image denoising techniques and the proposed CA, we used the Peak Signal to Noise Ratio (PSNR) \cite{Gonzalez06} and the Structural Similarity  Index Metric (SSIM) \cite{Wang04}. PSNR and SSIM metrics were calculated for the proposed method. To benchmark our results with the state-of-the-art, we used the PSNR and SSIM values reported in the literature for a variety of methods, namely, AMF \cite{Hwang95}, SMF \cite{Bovik00}, BDND \cite{Ng06}, MBUTMF \cite{Esakkirajan11}, Chan \textit{et al.} \cite{Chan05}, Sahin \textit{et al.} \cite{Sahin14}, DWMF \cite{Dong07}, MDWMF \cite{Lu12}, Zuoyong Li \textit{et al.} \cite{Li14}, PSMF \cite{Wang99}, IDBA \cite{Nair08}, Thirilogasundari \textit{et al.} \cite{Thirilogasundari12}, EDBA \cite{Srinivasan07}, FBAMF \cite{Punyaban12}, FBDA \cite{Nair10}, REBF \cite{Vijaykumar08}, DAMF \cite{Punyaban12b}, Pattnaik Ashutosh \textit{et al.} \cite{Pattnaik12} {for the same filtering window, i.e. $3\times{3}$}. To compare with the performance of the aforementioned methods, we used the PSNR and SSIM values reported in the literature.

In our experiments, the algorithms were implemented in Matlab R2014a on a laptop PC with Core i3 CPU at 2.2 GHz, 8 GB RAM, and Windows 7-64 bit operating system. A MATLAB implementation of the proposed algorithm can be found here\footnote{http://utopia.duth.gr/nmitiano/MATLAB/Denoising\_code.rar}. Tables \ref{Table1}-\ref{Table3} present a comparison of three widely used images with resolution of $256\times 256$ (Lena, Baboon, Barbara), so that our measurements can be easily compared to older experiments. Each image was corrupted by salt $\&$ pepper noise with varying noise density from $10\%$ to $90\%$ with incremental step $10\%$. The results of the proposed algorithm are the average of 100 independent runs of the method for each case. In Fig \ref{Fig7}, several denoising examples of the three $256\times 256$ images (Lena, Baboon, Barbara) are shown to facilitate objective evaluation. It can be seen that the proposed algorithm yields the highest PSNR and SSIM values among the other tested denoising methods. Larger values of PSNR indicate better quality of the restored image as well as larger SSIM value means that there is bigger structural similarity between the restored image and the original one. It is important that the proposed method outperforms previous offerings in lower resolution images, such as $256\times 256$, since it is well known that smaller images contain less spatial information, i.e. less detail around each examined pixel and the denoising task is much more difficult compared to higher resolutions thus resulting to blurred restored images.

Some of the other algorithms, such as SMF, PSMF, BDND, suffer from the blur effect in the restored image, producing unsatisfactory visual results. Nevertheless, some other algorithms, such as FBDA, Chan \textit{et al.} or DAMF, increase the quality of the restored image at a satisfying level.

\begin{figure*}
\centering
\includegraphics[scale=.3]{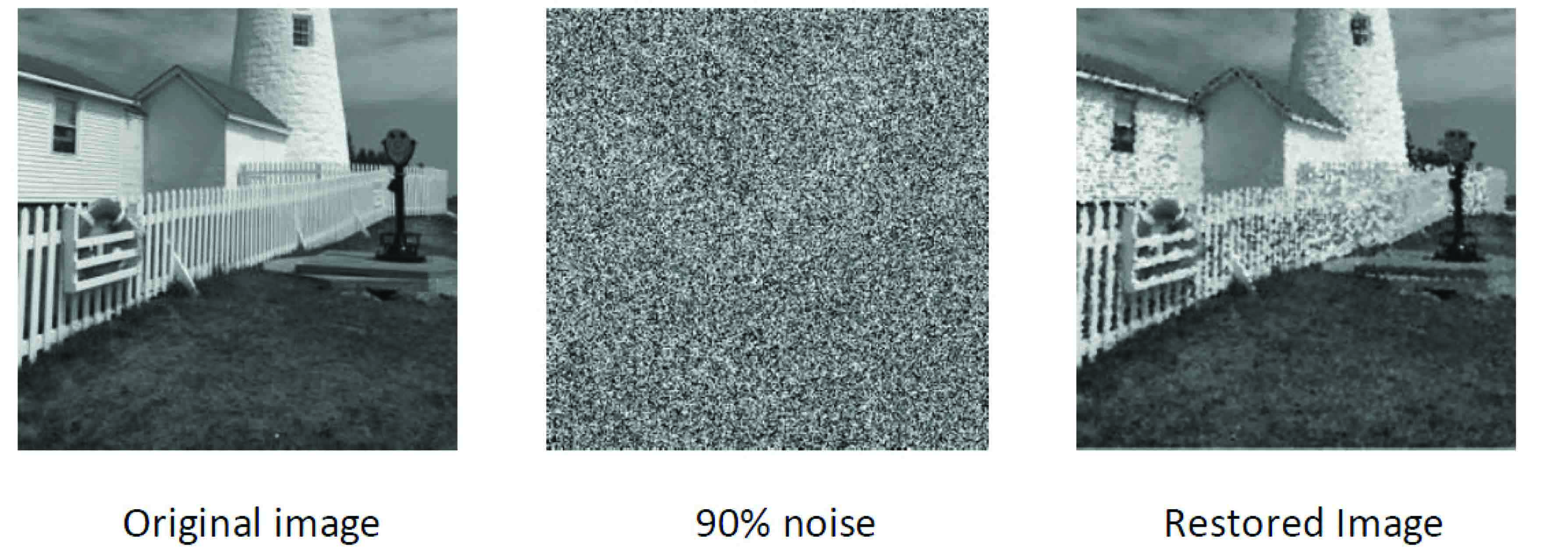}
\includegraphics[scale=.6]{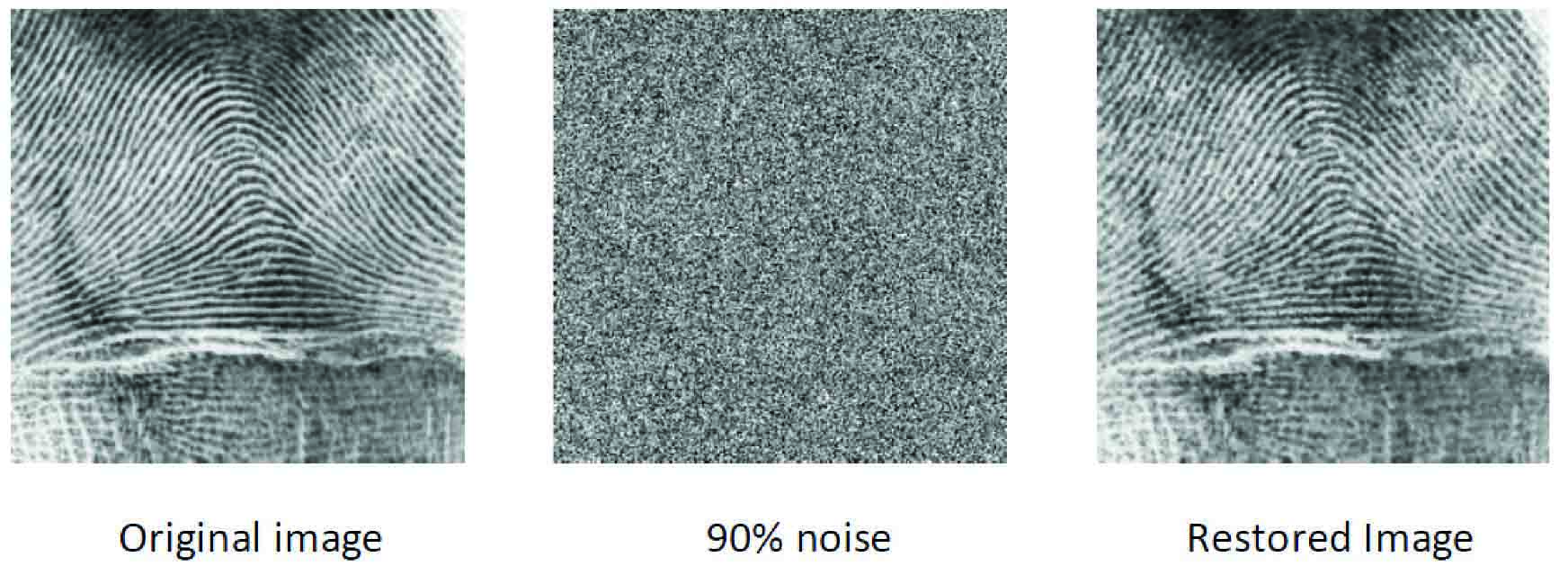}\\
\includegraphics[scale=.6]{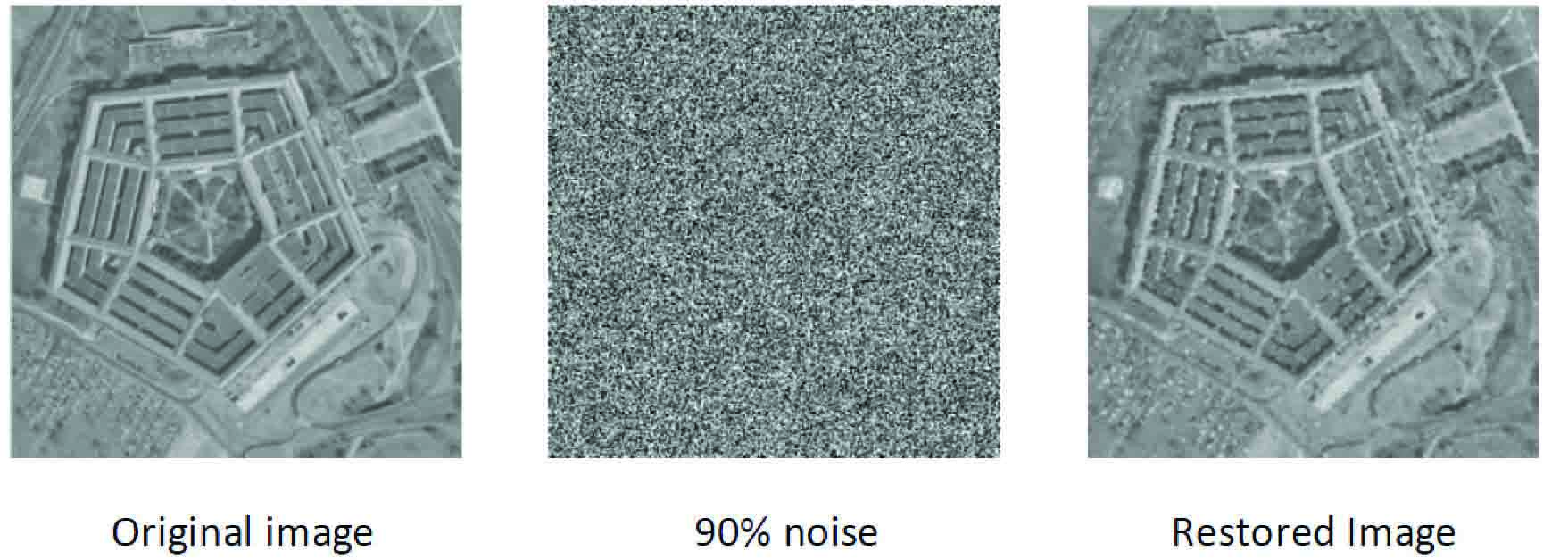}
\caption{Restored images using {the proposed filtering method} for $90\%$ of salt and pepper noise for lighthouse, fingerprint and pentagon $512\times 512$ pixel images.}
\label{Fig8}
\end{figure*}

Moreover, Table \ref{Table5} shows denoising examples of Lena at resolution $512\times 512$ for different noise ratios. Fig. \ref{Fig4} and Fig. \ref{Fig5} show denoising examples of Lena at resolution $512\times 512$ for $70\%$ and $90\%$ noise ratios {presenting in a qualitatively point of view, the application of numerous different filters to the same image and their results}. Again, the proposed method excels giving PSNR 27.98 dB at 90\% noise with the second method (FBAMF) giving 26.91 dB. At 70\%, the proposed method scores the highest score of 33.18 dB with the second method (FBAMF) giving 31.38 dB. In general, even if the proposed methods can be classified to low complexity and high complexity, like \cite{Chan05}, with the later ones extremely more demanding in computational sources \cite{Lien13}, the proposed low complexity method successfully outperforms all the methods described in literature, as already cited above. In Fig. \ref{Fig8}, it can also be observed that in high noise densities, such as $90\%$, the proposed method produces very satisfactory restoration results, considering the fact that much information is missing.

\section{Conclusions}
\label{Conclusions}
In this paper, a novel algorithm was proposed to eliminate the salt and pepper noise from  images using CA. The proposed algorithm was tested against different images and it yields excellent PSNR and SSIM values in comparison with existing methods. This method shows significant improvement, as it can remove the impulsive noise, varying from $10\%-90\%$, while keeping the blur of the image and the edges largely unaffected. To improve the filtering performance many different rules at different locations can be applied. Furthermore, due to the inherent parallelism of the proposed method, it can be easily implemented in any hardware parallel media, including Field-Programmable Gate Array (FPGA) and/or Graphics Processing Unit (GPU).

\bibliography{refsv1}

\begin{thebibliography}{10}

\bibitem{abu2011}
Abdel-Latif Abu-Dalhoum, I.~Al-Dhamari, Alfonso Ortega de~la Puente, and Manuel
  Alfonseca.
\newblock (2011).
\newblock Enhanced cellular automata for image noise removal.
\newblock In {\em Proceedings Asian Simulation Technology Conference ({ASTEC}
  2011)}.

\bibitem{Andreadis}
I.~Andreadis, P.~Iliades, Y.~Karafyllidis, Ph. Tsalides, and A.~Thanailakis.
\newblock (Jun 1995).
\newblock Design and {VLSI} implementation of a new {ASIC} for colour
  measurement.
\newblock {\em Circuits, Devices and Systems, IEE Proceedings -},
  142(3):153--157.

\bibitem{Bovik00}
A.~Bovik.
\newblock (2000).
\newblock {\em Handbook of Image and Video Processing}.
\newblock Academic Press.

\bibitem{Cappellari}
L.~Cappellari, S.~Milani, C.~Cruz-Reyes, and G.~Calvagno.
\newblock (May 2011).
\newblock Resolution scalable image coding with reversible cellular automata.
\newblock {\em Image Processing, IEEE Transactions on}, 20(5):1461--1468.

\bibitem{Chan05}
R.~H. Chan, C.W. Ho, and M.~Nikolova.
\newblock (2005).
\newblock Salt-and-pepper noise removal by median-type noise detectors and
  detail-preserving regularization.
\newblock {\em IEEE Trans. Image Process.}, 14(10):1479 -- 1485.

\bibitem{Chatzichristofis}
Savvas~A. Chatzichristofis, Dimitris~A. Mitzias, Georgios~Ch. Sirakoulis, and
  Yiannis~S. Boutalis.
\newblock (2010).
\newblock A novel cellular automata based technique for visual multimedia
  content encryption.
\newblock {\em Optics Communications}, 283(21):4250 -- 4260.

\bibitem{Chopard98}
B.~Chopard and M.~Droz.
\newblock (1998).
\newblock {\em Cellular Automata Modelling of Physical Systems}.
\newblock Cambridge Univ. Press, UK.

\bibitem{Dakua14}
Sarada~Prasad Dakua.
\newblock (January 2014).
\newblock Annularcut: a graph-cut design for left ventricle segmentation from
  magnetic resonance images.
\newblock {\em IET Image Processing}, 8:1--11(10).

\bibitem{DelRey}
A.~Martín Del~Rey and G.~Rodríguez Sánchez.
\newblock (2015).
\newblock An image encryption algorithm based on 3d cellular automata and
  chaotic maps.
\newblock {\em International Journal of Modern Physics C}, 26(01):1450069.

\bibitem{Dogaru}
R.~Dogaru, I.~Dogaru, and Hyongsuk Kim.
\newblock (Feb 2010).
\newblock Chaotic scan: A low complexity video transmission system for
  efficiently sending relevant image features.
\newblock {\em Circuits and Systems for Video Technology, IEEE Transactions
  on}, 20(2):317--321.

\bibitem{Dong07}
Y.~Q. Dong and S.~F. Xu.
\newblock (2007).
\newblock A new directional weighted median filter for removal of random-valued
  impulsive noise.
\newblock {\em IEEE Signal Process. Lett.}, 14(3):31--34.

\bibitem{Duff84}
J.~Duff and K.~Preston.
\newblock (1984).
\newblock {\em Modern Cellular Automata: Theory and Applications}.
\newblock Plenum Press.

\bibitem{Esakkirajan11}
S.~Esakkirajan, T.~Veerakumar, A.N. Subramanyam, and C.H. PremChand.
\newblock (2011).
\newblock Removal of high density salt and pepper noise through modified
  decision based unsymmetric trimmed median filter.
\newblock {\em IEEE Signal Process. Lett.}, 18(5):287--290.

\bibitem{Espinola}
M.~Espinola, J.A. Piedra-Fernandez, R.~Ayala, L.~Iribarne, and J.Z. Wang.
\newblock (Feb 2015).
\newblock Contextual and hierarchical classification of satellite images based
  on cellular automata.
\newblock {\em Geoscience and Remote Sensing, IEEE Transactions on},
  53(2):795--809.

\bibitem{Gao2011}
Yonghui Gao, Jie Yang, Xian Xu, and Feng Shi.
\newblock (2011).
\newblock Efficient cellular automaton segmentation supervised by pyramid on
  medical volumetric data and real time implementation with graphics processing
  unit.
\newblock {\em Expert Systems with Applications}, 38(6):6866 -- 6871.

\bibitem{Georgoulas2008}
Christos Georgoulas, Leonidas Kotoulas, Georgios~Ch. Sirakoulis, Ioannis
  Andreadis, and Antonios Gasteratos.
\newblock (2008).
\newblock Real-time disparity map computation module.
\newblock {\em Microprocessors and Microsystems}, 32(3):159 -- 170.

\bibitem{Gonzalez06}
Rafael~C. Gonzalez and Richard~E. Woods.
\newblock (2006).
\newblock {\em Digital Image Processing (3rd Edition)}.
\newblock Prentice-Hall, Inc., Upper Saddle River, NJ, USA.

\bibitem{Hamamci}
A.~Hamamci, N.~Kucuk, K.~Karaman, K.~Engin, and G.~Unal.
\newblock (March 2012).
\newblock Tumor-cut: Segmentation of brain tumors on contrast enhanced mr
  images for radiosurgery applications.
\newblock {\em Medical Imaging, IEEE Transactions on}, 31(3):790--804.

\bibitem{chinese}
Chih-Yu Hsu, Ta-Shan Tsui, Shyr-Shen Yu, and Kuo-Kun Tseng.
\newblock (09 2011).
\newblock Salt and pepper noise reduction by cellular automata.
\newblock {\em International Journal of Applied Science and Engineering},
  9:143--160.

\bibitem{Hwang95}
H.~Hwang and R.~A. Hadded.
\newblock (1995).
\newblock Adaptive median filter: New algorithms and results.
\newblock {\em IEEE Trans. Image Process.}, 4(4):449--502.

\bibitem{Ioannidis}
Konstantinos Ioannidis, Ioannis Andreadis, and GeorgiosCh. Sirakoulis.
\newblock (2012).
\newblock An edge preserving image resizing method based on cellular automata.
\newblock In GeorgiosCh. Sirakoulis and Stefania Bandini, editors, {\em
  Cellular Automata}, volume 7495 of {\em Lecture Notes in Computer Science},
  pages 375--384. Springer Berlin Heidelberg.

\bibitem{jana}
Biswapati Jana, Pabitra Pal, and Jaydeb Bhaumik.
\newblock (05 2012).
\newblock New image noise reduction schemes based on cellular automata.
\newblock {\em International Journal of Soft Computing and Engineering},
  2:98--103.

\bibitem{Jin2012}
J.~Jin.
\newblock (2012).
\newblock An image encryption based on elementary cellular automata.
\newblock {\em Optics and Lasers in Engineering}, 50(12):1836--1843.

\bibitem{Konstantinidis}
K.~Konstantinidis, A.~Amanatiadis, S.A. Chatzichristofis, R.~Sandaltzopoulos,
  and G.C. Sirakoulis.
\newblock (Oct 2014).
\newblock Identification and retrieval of dna genomes using binary image
  representations produced by cellular automata.
\newblock In {\em Imaging Systems and Techniques (IST), 2014 IEEE International
  Conference on}, pages 134--137.

\bibitem{Lafe00}
O.~Lafe.
\newblock (2000).
\newblock {\em Cellular Automata Transforms: Theory and Applications in
  Multimedia Compression. Encryption and Modeling}.
\newblock Kluwer Academic Publishers.

\bibitem{Li14}
Z.~Li, G.~Liu, Y.~Xu, and Y.~Cheng.
\newblock (2014).
\newblock Modified directional weighted filter for removal of salt and pepper
  noise.
\newblock {\em Pattern Recognition Letters}, 40:113--120.

\bibitem{Lien13}
Chih-Yuan Lien, Chien-Chuan Huang, Pei-Yin Chen, and Yi-Fan Lin.
\newblock (2013).
\newblock An efficient denoising architecture for removal of impulse noise in
  images.
\newblock {\em Computers, IEEE Transactions on}, 62(4):631--643.

\bibitem{Lu12}
C.~T. Lu and T.~C. Chou.
\newblock (2012).
\newblock Denoising of salt-and-pepper noise corrupted image using modified
  directional-weighted-median filter.
\newblock {\em Pattern Recognition Letters}, 13:1287--1295.

\bibitem{Mardiris08}
V.~Mardiris, G.~Ch. Sirakoulis, Ch. Mizas, I.~Karafyllidis, and A.~Thanailakis.
\newblock (2008).
\newblock A {CAD} system for modeling and simulation of computer networks using
  cellular automata.
\newblock {\em IEEE Trans. Systems, Man and Cybernetics-- Part C},
  38(2):253--264.

\bibitem{Hasanzadeh15}
Mohammad~Hasanzadeh Mofrad, Sana Sadeghi, Alireza Rezvanian, and Mohammad~Reza
  Meybodi.
\newblock (2015).
\newblock Cellular edge detection: Combining cellular automata and cellular
  learning automata.
\newblock {\em {AEU} - International Journal of Electronics and
  Communications}, 69(9):1282 -- 1290.

\bibitem{Nair10}
M.S. Nair and G.Raju.
\newblock (2010).
\newblock A new fuzzy-based decision algorithm for high-density impulse noise
  removal.
\newblock {\em Signal, Image and Video Processing}, 6:579--595.

\bibitem{Nair08}
M.S. Nair, K.~Revathy, and R.~Tatavarti.
\newblock (2008).
\newblock An improved decision-based algorithm for impulse noise removal.
\newblock In {\em Image and Signal Processing, 2008. CISP '08. Congress on},
  volume~1, pages 426--431.

\bibitem{Nalpantidis11more}
L.~Nalpantidis, A.~Amanatiadis, G.~Ch. Sirakoulis, and A.~Gasteratos.
\newblock (August 2011).
\newblock Efficient hierarchical matching algorithm for processing uncalibrated
  stereo vision images and its hardware architecture.
\newblock {\em IET Image Processing}, 5:481--492(11).

\bibitem{Nalpantidis11}
L.~Nalpantidis, G.~Ch. Sirakoulis, and A.~Gasteratos.
\newblock (2011).
\newblock Non-probabilistic cellular automata-enhanced stereo vision
  simultaneous localisation and mapping ({SLAM}).
\newblock {\em Measurement Science and Technology}, 22(11).

\bibitem{Ng06}
P.E. Ng and K.K. Ma.
\newblock (2006).
\newblock A switching median filter with boundary discriminative noise
  detection for extremely corrupted images.
\newblock {\em IEEE Trans. on Image Process.}, 15(6):1506--1516.

\bibitem{Pattnaik12}
A.~Pattnaik, S.~Agarwal, and S.~Chand.
\newblock (2012).
\newblock A new and efficient method for removal of high density salt and
  pepper noise through cascade decision based filtering algorithm.
\newblock {\em Procedia Technology}, 6:108--117.

\bibitem{Pitas90}
I.~Pitas and A.~N. Venetsanopoulos.
\newblock (1990).
\newblock {\em Nonlinear Digital Filters Principles and Applications}.
\newblock Kluwer, Norwell, MA.

\bibitem{Punyaban12b}
P.~Punyaban, M.~Banshidhar, J.~Bibekananda, and C.R. Tripathy.
\newblock (2012).
\newblock Dynamic adaptive median filter ({DAMF}) for removal of high density
  impulse noise.
\newblock {\em I.J. Image, Graphics and Signal Processing, Modern Education and
  Computer Science}, pages 53--62.

\bibitem{Punyaban12}
P.~Punyaban, M.~Banshidhar, J.~Bibekananda, and C.R. Tripathy.
\newblock (2012).
\newblock Fuzzy based adaptive median filtering technique for removal of
  impulse noise from images.
\newblock {\em Int. Jour. of Computer Vision and Signal Process.}, 1(1):15--21.

\bibitem{Rosin06}
P.~L. Rosin.
\newblock (2006).
\newblock Training cellular automata for image processing.
\newblock {\em IEEE Trans, Image Process.}, 15(7):2076--2087.

\bibitem{Rosin10}
P.~L. Rosin.
\newblock (2010).
\newblock Image processing using 3-state cellular automata.
\newblock {\em Computer Vision and Image Understanding}, 114(7):790--802.

\bibitem{Rosin14}
P.~L. Rosin, A.~Adamatzky, and X.~Sun.
\newblock (2014).
\newblock {\em Cellular Automata in Image Processing and Geometry}.
\newblock Springer.

\bibitem{Sadeghi12}
Sana Sadeghi, Alireza Rezvanian, and Ebrahim Kamrani.
\newblock (2012).
\newblock An efficient method for impulse noise reduction from images using
  fuzzy cellular automata.
\newblock {\em {AEU} - International Journal of Electronics and
  Communications}, 66(9):772 -- 779.

\bibitem{Sahin14}
U.~Sahin, S.~Uguz, and F.~Sahin.
\newblock (2014).
\newblock Salt and pepper noise filtering with fuzzy-cellular automata.
\newblock {\em Computers and electrical engineering}, 40:59--69.

\bibitem{wim}
P.~J. Selvapeter and Wim Hordijk.
\newblock (Dec 2009).
\newblock Cellular automata for image noise filtering.
\newblock In {\em 2009 World Congress on Nature Biologically Inspired Computing
  (NaBIC)}, pages 193--197.

\bibitem{Sirakoulis15}
G.~Ch. Sirakoulis and A.~Adamatzky.
\newblock (2015).
\newblock {\em Robots and Lattice Automata}.
\newblock Springer.

\bibitem{Srinivasan07}
K.~S. Srinivasan and D.~Ebenezer.
\newblock (2007).
\newblock A new fast and efficient decision-based algorithm for removal of
  high-density impulsive noises.
\newblock {\em IEEE Signal Proc. Letters}, 14(3):189--192.

\bibitem{Thirilogasundari12}
V.~Thirilogasundari, V.~Suresh babu, and S.~Agatha Janet.
\newblock (2012).
\newblock Fuzzy based salt and pepper noise removal using adaptive switching
  median filter.
\newblock {\em Procedia Engineering}, 38:2858--2865.

\bibitem{Uguz15}
S.~Uguz, U.~Sahin, and F.~Sahin.
\newblock (2015).
\newblock Edge detection with fuzzy cellular automata transition function
  optimized by \{PSO\}.
\newblock {\em Computers \& Electrical Engineering}, 43:180 -- 192.

\bibitem{Vijaykumar08}
V.R. Vijaykumar, P.T. Vanathi, P.~Kanagasabapathy, and D.Ebenezer.
\newblock (2008).
\newblock High density impulse noise removal using robust estimation based
  filter.
\newblock {\em Int. Jour. of Computer Science}, 35(3).

\bibitem{vonNeumann52}
J.~von Neumann.
\newblock (1952).
\newblock {\em Theory of automata}.
\newblock Urbana University Press.

\bibitem{Wang2013}
X.~Wang and D.~Luan.
\newblock (2013).
\newblock A novel image encryption algorithm using chaos and reversible
  cellular automata.
\newblock {\em Communications in Nonlinear Science and Numerical Simulation},
  18(11):3075--3085.

\bibitem{Wang04}
Z.~Wang, A.~C. Bovik, H.~R. Sheikh, and E.~P. Simoncelli.
\newblock (2004).
\newblock Image quality assessment: From error visibility to structural
  similarity.
\newblock {\em IEEE Trans. on Image Processing}, 13(4):600--612.

\bibitem{Wang99}
Z.~Wang and D.~Zhang.
\newblock (1999).
\newblock Progressive switching median filter for the removal of impulsive
  noise from highly corrupted images.
\newblock {\em IEEE Trans Circ Syst II, Analog Digit Signal Process},
  46(1):78--80.

\bibitem{Zhou12}
Y.Y. Zhou, Z.F. Ye, and J.J. Huang.
\newblock (2012).
\newblock Improved decision-based detail-preserving variational method for
  removal of random-valued impulse noise.
\newblock {\em IET Image Processing}, 6(7):976�985.

\end{thebibliography}
\bibliographystyle{ijuc}
\appendix

\end{document}